%% file: main.tex
\documentclass[journal]{IEEEtran}
\usepackage{amsmath,amsfonts}
\usepackage{algorithmic}
\usepackage{array}
\usepackage{textcomp}
\usepackage{stfloats}
\usepackage{url}
\usepackage{verbatim}
\usepackage{graphicx}
\usepackage[tight]{subfigure}
\def\BibTeX{{\rm B\kern-.05em{\sc i\kern-.025em b}\kern-.08em
    T\kern-.1667em\lower.7ex\hbox{E}\kern-.125emX}}
\usepackage{balance}

\usepackage{diagbox}
\usepackage{wrapfig}
\usepackage{algorithm}
\usepackage{algorithmic}
\usepackage{mathtools}
\usepackage{bm}
\usepackage{soul,color, xcolor}
\soulregister\cite7 
\soulregister\citep7 
\soulregister\citet7 
\soulregister\ref7 
\soulregister\pageref7 
\usepackage{booktabs}
\usepackage{bbding}
\usepackage{subcaption}
\usepackage{makecell,multirow}
\usepackage{colortbl}
\definecolor{mygray}{gray}{.9}
\definecolor{mypink}{rgb}{.99,.91,.95}
\definecolor{mycyan}{cmyk}{.3,0,0,0}
\usepackage[pagebackref=true,breaklinks=true, colorlinks,bookmarks=false]{hyperref}
\UseRawInputEncoding
\newcommand{\fakeparagraph}[1]{\vspace{1mm}\noindent\textbf{#1}}
\usepackage{cite}
\usepackage{pifont} %
\begin{document}
\title{Uncertainty-Guided Adverse Weather Restoration via Gated Transformer Network}
\author{Zheke Jin\,$^{2}$, Yuning Cui\,$^{2}$, Tianle Jin\,$^{2}$, Alois Knoll\,$^{2}$,~\IEEEmembership{Fellow,~IEEE}, Hu Cao\,$^{1*}$,~\IEEEmembership{Member,~IEEE}
\thanks{$^*$Hu Cao is the corresponding author of this work (hu.cao@seu.edu.cn).}
\thanks{Authors Affiliation: $^{1}$School of Automation, Southeast University, Nanjing, China,
$^{2}$Chair of Robotics, Artificial Intelligence and Real-time Systems, Technical University of Munich, Munich, Germany.}}

\markboth{Journal of \LaTeX\ Class Files,~Vol.~18, No.~9, September~2020}%
{How to Use the IEEEtran \LaTeX \ Templates}

\maketitle

\begin{abstract}
Restoring images degraded by adverse weather remains challenging due to spatially heterogeneous degradations.
Many existing weather-specific restoration models rely on weather-agnostic global aggregation, naive cross-scale fusion, and deterministic objectives, which struggle to handle heterogeneous degradations in all-in-one adverse-weather settings.
To address these limitations, we propose an \textbf{U}ncertainty-guided \textbf{A}dverse-weather \textbf{R}estoration \textbf{Net}work (UAR-Net), a weather-specific AiO framework that integrates a gated transformer with balanced multi-scale skip connections.
Specifically, we employ Gated Dual-scale Transformer Blocks (GDTB) to jointly model selective global interactions and multi-scale local structures, a progressive Balanced Multi-scale Skip Connection (BMSC) for balanced multi-scale feature integration, and an Uncertainty-Aware Refinement Head (URH) that performs artifact removal, detail enhancement, and predictive uncertainty estimation.
The model is supervised by a Brightness-Aware Energy Loss (BAE-Loss) to encourage accurate reconstruction with well-calibrated uncertainty.
Extensive experiments demonstrate that our method achieves state-of-the-art performance across multiple adverse-weather benchmarks. The codes will open source upon acceptance.
\end{abstract}


\input{Sections/1_Introduction}

\input{Sections/2_RelatedWorks}
\input{Sections/3_Method}

\input{Sections/4_Experiments}
\input{Sections/5_Conclusion}

\bibliographystyle{IEEEtran}
\bibliography{main}

\end{document}

%% file: Sections/1_Introduction.tex
\section{Introduction}

\IEEEPARstart{I}{mage} restoration aims to reconstruct high-quality images from degraded observations and is fundamental to modern computer vision. This is especially critical in safety-related applications such as autonomous driving~\cite{mușat2021multi,cao2021fusion,almalioglu2022deep} and intelligent surveillance~\cite{li2020all}, where rain, snow, and haze can obscure critical scene content, blur structural boundaries, and reduce visibility, thereby degrading the performance of downstream tasks like object detection, tracking~\cite{ren2015faster,carion2020end,hong2024you} and semantic segmentation~\cite{iqbal2022fogadapt}. Consequently, robust adverse-weather restoration has become a prerequisite for reliable perception in real-world systems rather than a purely aesthetic enhancement.

\begin{figure}[t]
    \centering
    \includegraphics[width=\columnwidth]{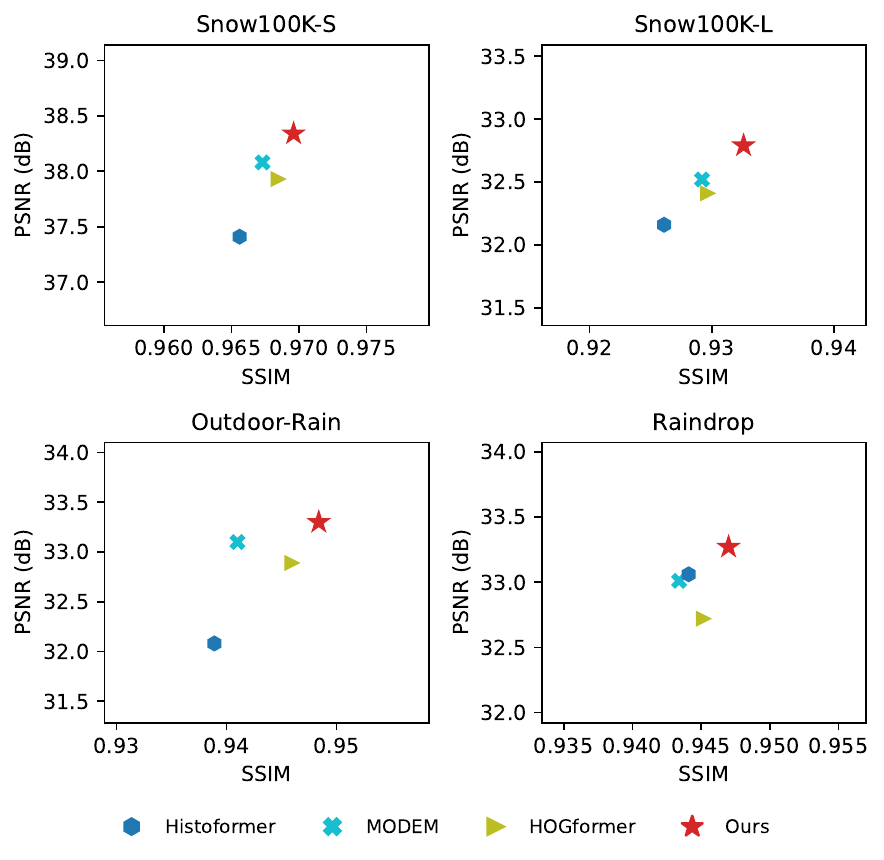}
    \caption{Quantitative comparison on four adverse-weather benchmarks,
    including Snow100K-S/L, Outdoor-Rain, and Raindrop. UAR-Net
    (\textbf{Ours}) achieves consistently strong PSNR and SSIM performance
    across all datasets, ranking favorably against recent SOTA methods and
    showing stable improvements under different adverse weather conditions.}
    \label{fig:intro_overall}
\end{figure}

Early adverse-weather restoration methods typically target a single degradation and rely on physical models or hand-crafted priors~\cite{he2010single,ancuti2013single,berman2016non}, which often struggle in complex or real-world scenarios. With deep learning, data-driven models have significantly improved performance for specific tasks such as deraining~\cite{chen2018robust,jiang2020multi}, dehazing~\cite{li2019semi,wu2021contrastive}, and desnowing~\cite{zhang2021deep}. However, these methods are designed for individual weather conditions, requiring separate models for different degradations, which limits scalability and robustness in unconstrained environments.
Recent advances in adverse-weather image restoration have shifted from task-specific models to unified All-in-One (AiO) frameworks tailored for multiple weather degradations within a single network~\cite{chen2022simple,mou2022deep}. Early AiO methods rely on shared representations but have limited capacity for global context modeling. More recent transformer-based AiO approaches leverage hierarchical architectures to capture multi-scale features and long-range dependencies, achieving improved restoration performance across diverse weather conditions~\cite{sun2024restoring,zhu2024mwformer,wang2025modem}. However, most existing approaches aggregate global context in a uniform and weather-agnostic manner, even though different degradations rely on global information differently.
This often leads to suboptimal context representations and weak emphasis on severely degraded regions.
Meanwhile, skip connections are often implemented as simple concatenation or addition, which may directly propagate heavily corrupted and variable shallow features across weather conditions.
This can amplify degradation-specific noise and weaken effective cross-scale feature interaction, hindering robust restoration.
Finally, most AiO models adopt deterministic objectives with point-wise predictions.
In a unified setting with diverse weather conditions and degradation levels,
ignoring predictive uncertainty often leads to over-confident and unstable restorations in ambiguous or severely degraded regions.


To address these issues, we propose UAR-Net, a unified adverse-weather restoration network that integrates GDTB with BMSC and URH.
Each GDTB combines Selective Gated Attention (SGA) and a Dual-Scale Gated Feed-Forward (DGFF)~\cite{sun2024restoring}, where SGA applies sinusoidal reweighting to queries and keys together with data-dependent gating to selectively emphasize degradation-related long-range interactions, while DGFF provides complementary local enhancement.
Beyond the backbone, we introduce BMSC, which progressively integrates multi-level encoder features instead of performing one-shot concatenation or addition.
This design jointly considers multi-scale information~\cite{pang2019libra,cao2024sdpt}. By controlling cross-scale information, BMSC reduces the impact of corrupted shallow features and provides more balanced guidance for stable decoding.
Building upon the coarse prediction, we introduce URH, which performs fine-grained correction while explicitly estimating pixel-wise uncertainty.
The refinement is supervised by the proposed BAE-Loss,
leading to robust restoration in severely degraded regions. As shown in Fig.~\ref{fig:intro_overall}, extensive experiments demonstrate that our proposed method achieves SOTA performance across multiple benchmarks, consistently outperforming recent unified models ~\cite{sun2024restoring,wang2025modem,wu2025beyond}.
The main contributions of this work are summarized below:
\begin{itemize}
\item We propose UAR-Net, which adopts GDTB with query-conditioned gating to selectively regulate global context aggregation, enabling weather-aware and adaptive feature modeling under diverse adverse conditions.

\item We introduce BMSC that replaces direct skip concatenation with progressive cross-scale integration, producing cleaner and better-balanced skip representations and enabling more stable and effective cross-scale information transfer for decoding.

\item We propose URH together with BAE-Loss to refine visual details and explicitly model pixel-wise uncertainty, resulting in sharper reconstructions and more reliable uncertainty estimates under severe degradations.

\item Extensive experiments on multiple adverse-weather benchmarks demonstrate that our method achieves SOTA performance compared with recent unified restoration models.
\end{itemize}

%% file: Sections/2_RelatedWorks.tex
\section{Related Work}

This section reviews representative approaches to adverse-weather image restoration, covering both task-specific methods and unified AiO frameworks.

\subsection{Task-specific Adverse-Weather Restoration}
Early deep learning approaches mainly focus on restoring images degraded by a single type of adverse weather, such as rain streaks, snow particles, or raindrops.
For rain streak removal, representative methods aim to separate rain structures from background textures using high-frequency decomposition~\cite{fu2017removing} and recurrent context aggregation~\cite{li2018recurrent}.
Subsequent works improve robustness through uncertainty-guided learning~\cite{yasarla2019uncertainty}.
More recently, transformer-based architectures have been introduced to better capture long-range dependencies in deraining~\cite{chen2023learning}.
For snow removal, existing methods therefore often incorporate explicit snow modeling~\cite{liu2018desnownet} or dense multi-scale architectures~\cite{li2019stacked}, with additional semantic or contextual priors to improve robustness under heavy snow conditions~\cite{zhang2021deep}.
For raindrop removal, representative approaches formulate the task as attention-guided image-to-image translation~\cite{qian2018attentive} or adopt strong residual learning frameworks for robust inpainting~\cite{liu2019dual}, while recent models further enhance contextual reasoning using transformer-based designs~\cite{tu2022maxim}.

\subsection{All-in-One Adverse-Weather Restoration}
AiO restoration seeks to address multiple adverse-weather degradations within a single unified framework.
Early AiO approaches employ recurrent architectures to jointly handle different degradations such as rain and haze~\cite{li2018recurrent}, or explore architecture-search-based designs for multi-task restoration~\cite{li2020all}.
Recent advances are largely driven by transformer-based models, which leverage hierarchical architectures and global self-attention to model diverse degradations more effectively.
Representative works include histogram-based self-attention for intensity-aware restoration~\cite{sun2024restoring}, degradation-aware transformers~\cite{zhu2024mwformer}, gradient-conditioned attention with explicit priors~\cite{wu2025beyond}, and Morton-order scanning with dual degradation estimation~\cite{wang2025modem}.
However, in the AiO setting existing approaches still face notable challenges. Global context is often aggregated in a uniform manner, even though different weather conditions rely on long-range information in distinct ways.
And multi-scale feature fusion remains unreliable, as shallow representations exhibit highly variable reliability across adverse-weather scenarios.


%% file: Sections/3_Method.tex
\section{Method}
\begin{figure*}[t]
    \centering
    \includegraphics[width=\linewidth]{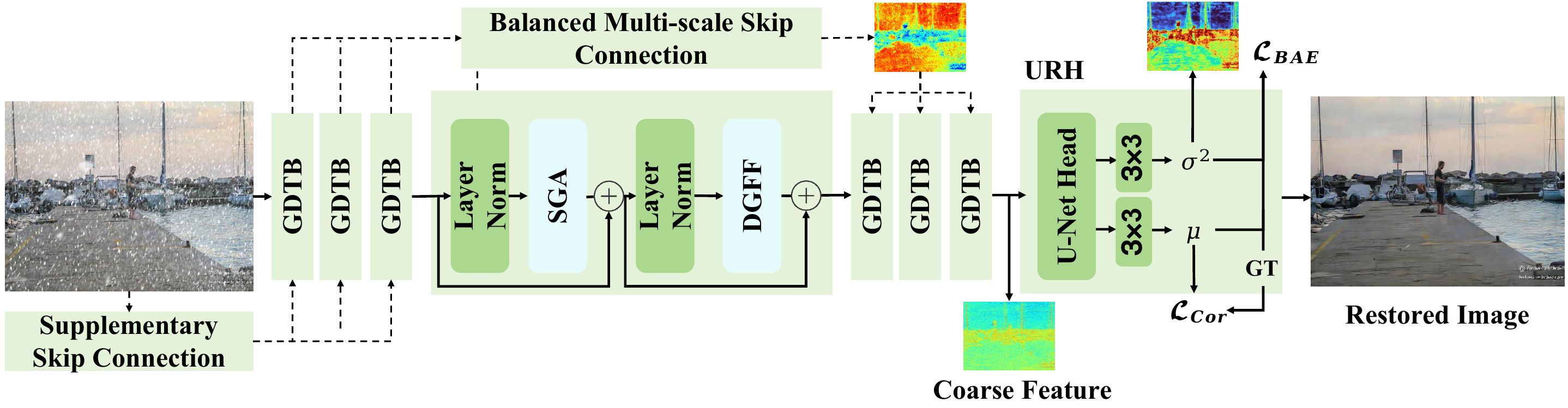}
    \caption{
    Overall architecture of the proposed \textbf{UAR-Net}.
The network follows a U-Net–style encoder–decoder design built upon
GDTB. Balanced Multi-scale Skip Connection (BMSC) progressively integrates multi-level
encoder features to form a balanced skip representation.
In addition, a Supplementary Skip connection composed of average pooling,
$1\times1$ convolution, and $3\times3$ depth-wise convolution injects
low-level structural cues with minimal computational overhead.
The decoder produces a coarse feature representation that is further refined by
URH.
Through parallel mean and variance prediction heads, the network outputs the
final restored image together with a pixel-wise uncertainty map, which is
supervised by the proposed BAE-Loss and a
Correlation Loss to enforce structural
consistency.
    }
    \label{fig:framework_umgt}
\vspace{-10pt}
\end{figure*}

\subsection{Overview}
As shown in Fig.~\ref{fig:framework_umgt}, UAR-Net is a Transformer-based
framework for adverse-weather image restoration.
Both the encoder and decoder are built from stacked GDTBs for hierarchical feature extraction and reconstruction.
Each GDTB integrates SGA, which introduces query-conditioned gating into linear attention to selectively regulate long-range context aggregation, followed by a dual-scale gated feed-forward module for local detail enhancement. To facilitate effective cross-scale interaction, BMSC progressively aggregates features from multiple encoder stages into a balanced skip representation. To retain low-frequency priors
and facilitate residual learning, the supplementary skip connections~\cite{sun2024restoring} are introduced to highlight degradation residuals through average pooling, pointwise convolution and depthwise convolution.
Finally, URH refines the coarse output and jointly predicts the restored image and uncertainty, supervised by the proposed BAE-Loss.

\subsection{Gated Dual-Scale Transformer Blocks (GDTB)}


\begin{figure}
\vspace{-10pt}
\centering
\includegraphics[width=\linewidth]{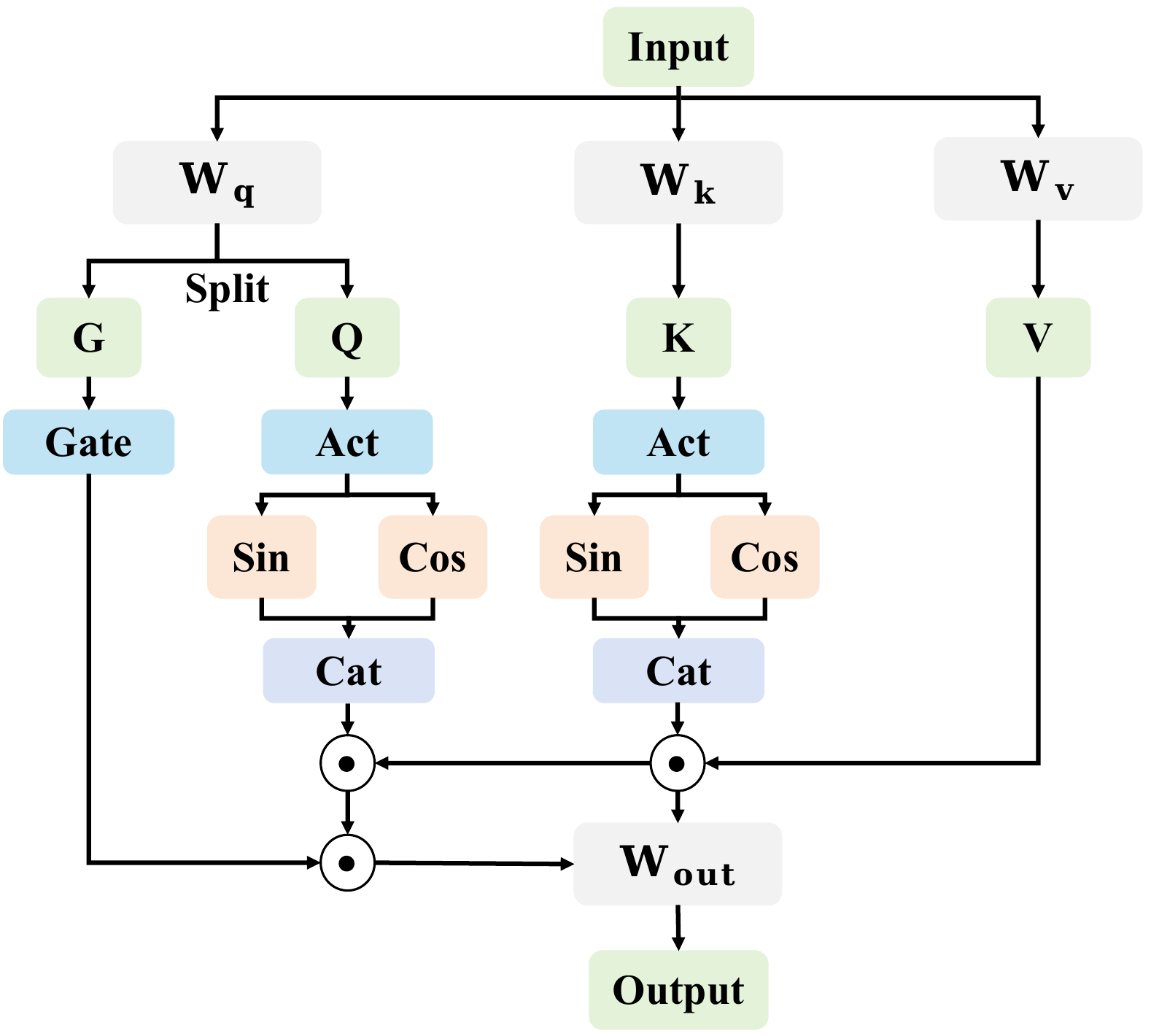}
\vspace{-6pt}
\caption{
Structure of the Selective Gated Attention (SGA).
The query projection is split into an attention query and a gating branch.
The gating branch applies a sigmoid activation to generate a data-dependent gate, which multiplicatively modulates the linear attention output, enabling selective and content-adaptive global context aggregation.
}
\label{fig:sgca}
\vspace{-10pt}
\end{figure}

The query projection is split into an attention query and a gating branch.
The gating branch applies a sigmoid activation to generate a data-dependent gate, which multiplicatively modulates the linear attention output, enabling selective and content-adaptive global context aggregation.

\paragraph{From Softmax Attention to Linear Attention}
Given query, key, and value matrices $Q,K,V\in\mathbb{R}^{N\times d}$, standard self-attention is defined as
\begin{equation}
\text{Attn}(Q,K,V)=\text{Softmax}\!\left(\frac{QK^\top}{\sqrt{d}}\right)V,
\end{equation}
which explicitly constructs the attention matrix in $\mathbb{R}^{N\times N}$ and therefore incurs $\mathcal{O}(N^2)$ time and memory complexity, becoming prohibitive for high-resolution inputs.
The attention mechanism was generalized by allowing arbitrary similarity functions between queries and keys~\cite{katharopoulos2020transformers}:
\begin{equation}
\text{Attention}(Q,K,V)_i = 
\frac{\sum_{j=1}^{N} \text{sim}(Q_i, K_j)V_j}
{\sum_{j=1}^{N} \text{sim}(Q_i, K_j)},
\label{eq:general_attention}
\end{equation}
where $\text{sim}(\cdot,\cdot)$ denotes a customizable similarity function.
When choosing
\[
\text{sim}(q,k)=\exp\!\left(\frac{q^\top k}{\sqrt{d}}\right),
\]
Eq.~\eqref{eq:general_attention} reduces to conventional softmax attention.
To obtain a decomposable similarity function, one can adopt a kernel
$\omega:\mathbb{R}^d\times\mathbb{R}^d\rightarrow\mathbb{R}^+$ that admits a feature map
$\phi(\cdot)$ such that
\begin{equation}
\text{sim}(x,y)=\omega(x,y)=\phi(x)^\top\phi(y).
\label{eq:kernel_sim}
\end{equation}
Substituting Eq.~\eqref{eq:kernel_sim} into Eq.~\eqref{eq:general_attention} yields
\begin{equation}
\text{Attention}(Q,K,V)_i = 
\frac{\sum_{j=1}^{N} \phi(Q_i)^\top \phi(K_j)V_j}
{\sum_{j=1}^{N} \phi(Q_i)^\top \phi(K_j)}.
\label{eq:kernel_attention}
\end{equation}
By exploiting the distributive and associative properties of matrix multiplication,
Eq.~\eqref{eq:kernel_attention} can be rewritten as
\begin{equation}
\text{Attention}(Q,K,V)_i =
\frac{\phi(Q_i)^\top \Big(\sum_{j=1}^{N} \phi(K_j)V_j \Big)}
{\phi(Q_i)^\top \Big(\sum_{j=1}^{N} \phi(K_j) \Big)}.
\label{eq:linear_attention}
\end{equation}
This reformulation avoids explicit computation of all pairwise dot-products $Q_i^\top K_j$.
Instead, it relies on two global summaries,
$\sum_{j=1}^{N}\phi(K_j)V_j$ and $\sum_{j=1}^{N}\phi(K_j)$,
which can be computed once and shared across all queries.
As a result, both time and memory complexity are reduced from $\mathcal{O}(N^2)$ to $\mathcal{O}(N)$ for a fixed head dimension.
In practice, Linear Attention commonly adopts
\[
\phi(x)=\text{ELU}(x)+1
\]
to ensure non-negativity and stable optimization.

\paragraph{Selective Gated Attention}

In image restoration, self-attention is computationally expensive on high-resolution feature maps, while Window-based or sparse variants limit global context modeling.
Linear attention demonstrates promise in global context modeling while maintaining linear complexity~\cite{ai2025breaking}.
However the AiO setting requires a single model to handle multiple weather degradations,
both self-attention and linear attention tend to aggregate global information uniformly. Such uniform aggregation can mix different degradation patterns and weaken degradation-related cues, limiting the model’s ability to adapt to different weather conditions.
Motivated by this observation, we adopt Selective Gated Attention (SGA) as the core attention mechanism in GDTB, as shown in Fig.~\ref{fig:framework_umgt}.
We build SGA on a sinusoidal feature mapping applied to queries and keys~\cite{qin2022cosformer}, where nearby or similar tokens receive higher
responses, implicitly encouraging locality in global aggregation. 
This property is particularly beneficial under different weather conditions, as it helps preserve locally coherent structures (e.g., rain streaks or snowflakes) while preventing distant and unrelated regions from being mixed together, enabling more adaptive use of global context across diverse weather scenarios.

To further enhance the reweighting capability of linear attention, we adopt
a sinusoidal modulation of query and key features~\cite{qin2022cosformer}.
Specifically, the similarity between a query at position $i$ and a key at position $j$
is modulated by a cosine function of their relative positions:
\begin{equation}
s(Q_i, K_j) = Q'_i (K'_j)^\top
\cos\!\left(\frac{\pi}{2}\cdot \frac{i-j}{M}\right),
\label{eq:sinusoidal_sim}
\end{equation}
where $Q'=\mathrm{ReLU}(Q)$, $K'=\mathrm{ReLU}(K)$, and $M \ge N$ is a normalization constant.
Using the trigonometric identity
$\cos(a-b)=\cos a \cos b + \sin a \sin b$,
Eq.~\eqref{eq:sinusoidal_sim} can be decomposed as
\begin{equation}
\begin{aligned}
Q'_i (K'_j)^\top
\cos\!\left(
\frac{\pi(i-j)}{2M}
\right)
={}&
\left(
Q'_i\cos\frac{\pi i}{2M}
\right)
\left(
K'_j\cos\frac{\pi j}{2M}
\right)^\top
\\
&+
\left(
Q'_i\sin\frac{\pi i}{2M}
\right)
\left(
K'_j\sin\frac{\pi j}{2M}
\right)^\top .
\end{aligned}
\end{equation}
Accordingly, we define the sinusoidally reweighted queries and keys as
\[
Q_i^{\cos} = Q'_i \cos\!\left(\tfrac{\pi i}{2M}\right), \quad
Q_i^{\sin} = Q'_i \sin\!\left(\tfrac{\pi i}{2M}\right),
\]
\[
K_j^{\cos} = K'_j \cos\!\left(\tfrac{\pi j}{2M}\right), \quad
K_j^{\sin} = K'_j \sin\!\left(\tfrac{\pi j}{2M}\right).
\]
With these definitions, the attention output at position $i$ can be expressed as
\begin{equation}
O_i =
\frac{
\sum_{j=1}^{N} Q_i^{\cos}(K_j^{\cos})^\top V_j
+
\sum_{j=1}^{N} Q_i^{\sin}(K_j^{\sin})^\top V_j
}{
\sum_{j=1}^{N} Q_i^{\cos}(K_j^{\cos})^\top
+
\sum_{j=1}^{N} Q_i^{\sin}(K_j^{\sin})^\top
}.
\label{eq:final_attention}
\end{equation}

This formulation preserves the linear computational complexity of kernelized attention
while introducing a structured reweighting mechanism through sinusoidal modulation.
The sine and cosine components act as complementary channels that encode relative positional
relationships and selectively reweight long-range interactions, thereby enhancing
expressiveness without sacrificing efficiency.

The attention output at position $i$ is given by
\begin{equation}
O_i =
\frac{\sum_{j=1}^{N} Q_i^{\cos} (K_j^{\cos})^\top V_j}
     {\sum_{j=1}^{N} Q_i^{\cos} (K_j^{\cos})^\top}
+
\frac{\sum_{j=1}^{N} Q_i^{\sin} (K_j^{\sin})^\top V_j}
     {\sum_{j=1}^{N} Q_i^{\sin} (K_j^{\sin})^\top},
\label{eq:cosformer_final}
\end{equation}
where $N$ denotes the number of tokens, $Q,K,V\in\mathbb{R}^{N\times d}$ are the query, key, and value matrices with head dimension $d$,
and $Q_i^{\cos},Q_i^{\sin}\in\mathbb{R}^{1\times d}$ (resp.\ $K_j^{\cos},K_j^{\sin}\in\mathbb{R}^{1\times d}$) denote the sinusoidally reweighted query and key features.
By avoiding explicit construction of the $N\times N$ attention matrix, the computational complexity is reduced from $\mathcal{O}(N^2)$ to $\mathcal{O}(N)$.

To enable more selective utilization of global context, SGA further incorporates a gating mechanism.
As illustrated in Fig.~\ref{fig:sgca}, the query features are projected and expanded to jointly generate
both attention queries and gating signals.
Formally, given the input feature $X_i$ at position $i$, we compute
\begin{equation}
  [\,Q_i \;\; G_i\,] = X_i W_q, \qquad
  K_i = X_i W_k, \qquad
  V_i = X_i W_v,
  \label{eq:qkv_split}
\end{equation}
where $W_q \in \mathbb{R}^{d \times 2d}$ and $W_k, W_v \in \mathbb{R}^{d \times d}$ are learnable projection matrices.
The expanded query projection is split channel-wise into two equal parts:
$Q_i \in \mathbb{R}^{d}$ is used for attention computation, while $G_i \in \mathbb{R}^{d}$ serves as a gating signal.
The attention output is then modulated as
\begin{equation}
  \tilde{O}_i = \sigma(G_i) \odot O_i,
  \label{eq:sgca}
\end{equation}
where $\sigma(\cdot)$ denotes the sigmoid function and $\odot$ represents element-wise multiplication. This gating introduces data-dependent non-linearity into the attention pathway~\cite{qiu2025gated}, breaking the purely linear transformation of the value and output projections and thereby enhancing the expressive capacity of attention modeling.
The gate also acts as a selective filter, suppressing less informative tokens while emphasizing relevant ones, 
leading to more focused global context modeling for diverse adverse-weather degradations.
Together, this leads to more focused global context modeling, which is beneficial for diverse adverse-weather degradations.

\paragraph{Dual-scale Gated Feed-Forward}

\begin{figure}
    \centering
    \includegraphics[width=0.9\linewidth]{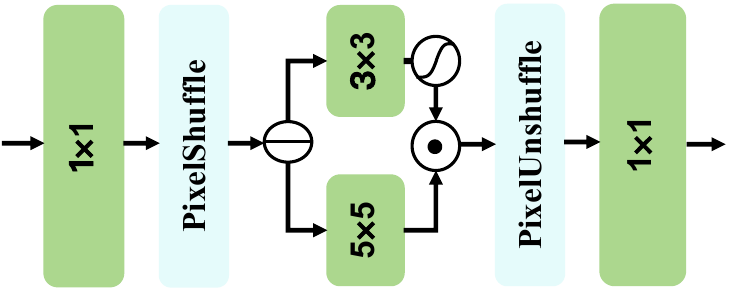}
    \caption{
    Structure of DGFF.
    DGFF applies a $1\times1$ projection, expands features via PixelShuffle, and uses two depth-wise convolution branches ($3\times3$ and $5\times5$) to capture local patterns at different receptive fields.
    A gating operation adaptively fuses the two branches, followed by PixelUnshuffle and a final $1\times1$ projection to produce the output.
    }
    \label{fig:dgff}
\end{figure}

To enhance local feature modeling under spatially heterogeneous adverse-weather degradations, GDTB adopts DGFF~\cite{sun2024restoring} by employing parallel depth-wise convolution branches with different receptive fields. Fig.~\ref{fig:dgff} illustrates the detailed structure of the Dual-Scale Gated Feed-Forward (DGFF) module.
DGFF enhances local feature modeling by capturing complementary spatial patterns at different receptive fields and adaptively fusing them through a gating mechanism, providing effective multi-scale local refinement within each GDTB.


\subsection{Balanced Multi-Scale Skip Connection (BMSC)}

\begin{figure}[t]
  \centering
  \includegraphics[width=\linewidth]{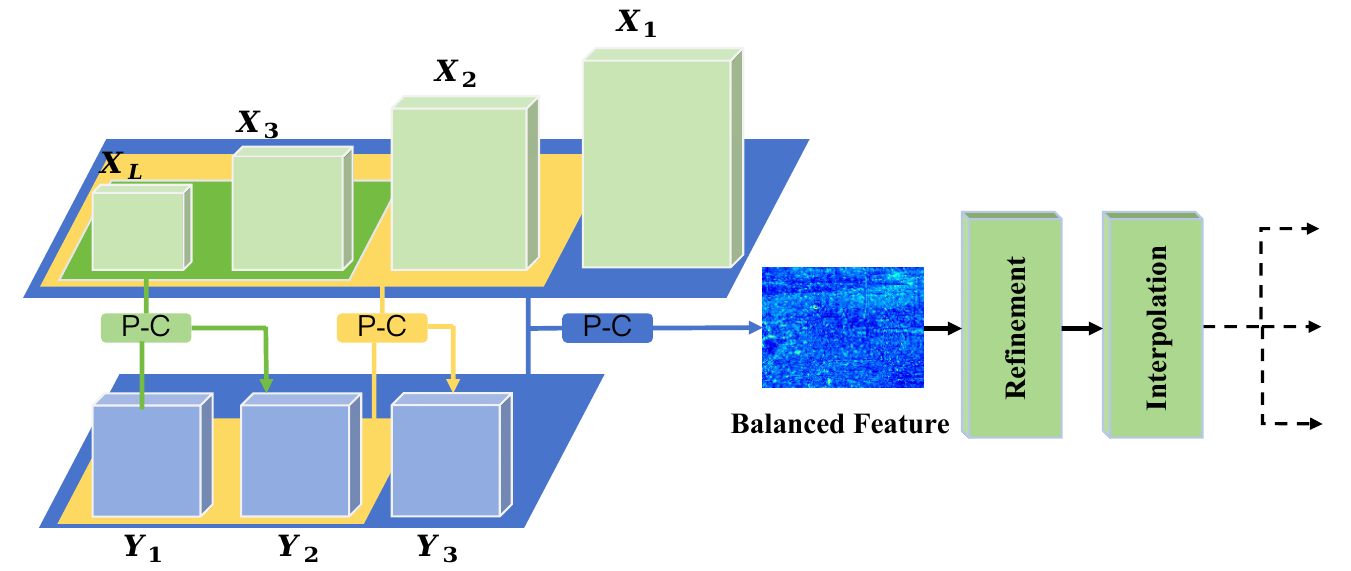}
  \caption{
Illustration of the proposed Balanced Multi-Scale Skip Connection (BMSC). Encoder features from multiple levels are progressively integrated via a predictor--corrector (P--C) scheme to form a balanced skip representation, which is lightly refined before being fed to the decoder as stable cross-scale guidance.
  }
  \label{fig:bms_pc}
\end{figure}

Instead of directly adding or concatenating encoder features to decoder features at the same resolution,
we introduce BMSC as a more robust skip pathway for adverse-weather image restoration.
BMSC adopts a top-down multi-scale fusion strategy inspired by FPN, progressively integrating
encoder features from deep to shallow into a single balanced skip representation.
This design is particularly important under diverse weather conditions, where shallow features
may be unevenly corrupted by different degradations (e.g., rain streaks, snowflakes, or haze).
By explicitly controlling cross-scale aggregation, BMSC suppresses unreliable shallow responses
while preserving complementary fine details and high-level semantics from deeper layers.
As illustrated in Fig.~\ref{fig:bms_pc}, the integration is implemented using a
predictor--corrector (P--C) scheme~\cite{he2025fuseunet} followed by a lightweight refinement,
providing clean and stable skip guidance for the decoder across different weather scenarios.

\paragraph{Multi-scale Integration}

\begin{table*}
  \centering
  \caption{Classical linear multistep methods (Adams--Bashforth and Adams--Moulton)}
  \label{tab:linear_multistep}
  \vskip 0.05in
  \small
  \setlength{\tabcolsep}{10pt}
  \renewcommand{\arraystretch}{1.2}
  \resizebox{\columnwidth}{!}{%
  \begin{tabular*}{\columnwidth}{@{\extracolsep{\fill}}ccc@{}}
    \toprule
    \multicolumn{3}{c}{\textbf{Explicit: Adams--Bashforth (AB)}} \\
    \midrule
    Step & Order & Equation \\
    \midrule
    1 & 1 &
    $y_{n+1} = y_n + \delta F_n$ \\
    2 & 2 &
    $y_{n+2} = y_{n+1} + \frac{\delta}{2}(3F_{n+1} - F_n)$ \\
    3 & 3 &
    $y_{n+3} = y_{n+2} + \frac{\delta}{12}(23F_{n+2} - 16F_{n+1} + 5F_n)$ \\
    4 & 4 &
    $y_{n+4} = y_{n+3} + \frac{\delta}{24}(55F_{n+3} - 59F_{n+2} + 37F_{n+1} - 9F_n)$ \\
    \midrule
    \multicolumn{3}{c}{\textbf{Implicit: Adams--Moulton (AM)}} \\
    \midrule
    Step & Order & Equation \\
    \midrule
    1 & 2 &
    $y_{n+1} = y_n + \frac{\delta}{2}(F_n + F_{n+1})$ \\
    2 & 3 &
    $y_{n+2} = y_{n+1} + \frac{\delta}{12}(5F_{n+2} + 8F_{n+1} - F_n)$ \\
    3 & 4 &
    $y_{n+3} = y_{n+2} + \frac{\delta}{24}(9F_{n+3} + 19F_{n+2} - 5F_{n+1} + F_n)$ \\
    \bottomrule
  \end{tabular*}%
  }
  \vskip -0.05in
\end{table*}

\begin{figure}[t]
  \centering
  \includegraphics[width=\linewidth]{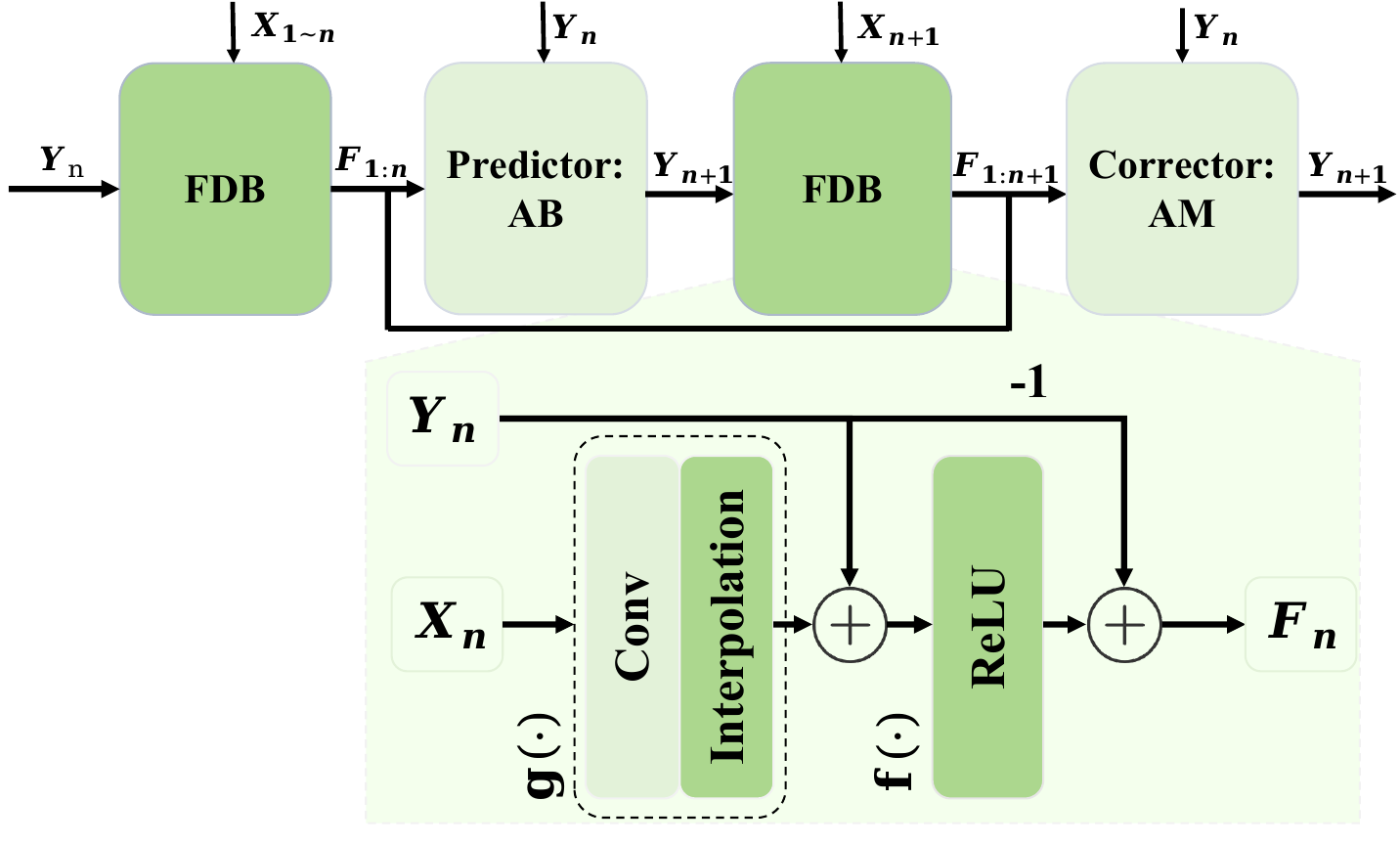}
  \caption{
Structure of the predictor--corrector (P--C) module and the Fusion Direction Block (FDB). The alignment function $g(\cdot)$ aligns features via convolution and interpolation, while the fusion function $f(\cdot)$ is implemented as element-wise addition followed by a $\mathrm{ReLU}(\cdot)$.
  }
  \label{fig:fdb}
\end{figure}

We view the progressive skip fusion process from the perspective of numerical integration.
Table~\ref{tab:linear_multistep} summarizes the explicit
Adams--Bashforth (AB) and implicit Adams--Moulton (AM)
schemes~\cite{he2025fuseunet}, which motivate the predictor--corrector update used
in our BMSC. When the number of integration steps is fixed, implicit methods generally provide higher accuracy and better numerical stability than explicit ones.
However, implicit schemes require the fusion direction at the current step, which is unknown before completing the update.
The P--C strategy provides a practical compromise: it first predicts the next state using an explicit scheme and then refines it using an implicit correction.
Let $\{X_n\}_{n=1}^{L}$ denote encoder features extracted at different depths,
where $n=1$ corresponds to the shallowest level and $n=L$ to the deepest (latent) level.
BMSC integrates these features from deep to shallow in a fixed order and progressively
constructs a balanced skip feature $\{Y_n\}$, where $n$ indexes the integration step.
At each step, the encoder feature is aligned to the current skip representation through
an operator $g(\cdot)$ composed of convolution and interpolation, ensuring matched spatial
resolution and channel dimension.
To enable a principled integration, we model the evolution of the balanced skip feature
as a continuous-time dynamical system:
\begin{equation}
\dot{Y}(t) = F\big(t, Y(t)\big)
= f\big(Y(t) + g(X(t))\big) - Y(t),
\label{eq:bms_continuous}
\end{equation}
where $f(\cdot)$ denotes a lightweight fusion operator implemented as element-wise addition
followed by a $\mathrm{ReLU}(\cdot)$ activation.
The decay term $-Y(t)$ serves as a stabilizing mechanism that prevents uncontrolled
accumulation of corrupted information, which is particularly important when integrating
noisy shallow features.
We use the continuous formulation as a conceptual model to explain how the balanced
skip feature evolves during multi-scale fusion.
In implementation, this evolution is realized by a sequence of discrete fusion steps.
At the $n$-th step, the fusion direction is evaluated as $F(Y_n, X_n)$.
Using a step size $\delta$, an explicit Euler discretization provides a first prediction:
\begin{equation}
\bar{Y}_{n+1} = Y_n + \delta \cdot F(Y_n, X_n),
\label{eq:bms_euler}
\end{equation}
where $\bar{Y}_{n+1}$ denotes the predicted balanced skip feature.
To improve robustness under severe adverse-weather corruption, we further apply a P--C update:
\begin{equation}
Y_{n+1}
=
Y_n + \frac{\delta}{2}
\Big(
F(Y_n, X_n) + F(\bar{Y}_{n+1}, X_{n+1})
\Big),
\label{eq:bms_pc}
\end{equation}
which can be viewed as a second-order Adams--Moulton correction.
This two-point update reduces sensitivity to noisy updates and promotes smoother cross-scale information propagation.
More generally, BMSC can be interpreted as a linear multistep integration scheme:
\begin{equation}
Y_{n+1} = Y_n + \delta \sum_{j=0}^{K-1} \beta_j \, F(Y_{n-j}, X_{n-j}),
\label{eq:bms_multistep}
\end{equation}
Here, $K$ denotes the number of integration steps and ${\beta_j}$ are fixed multistep coefficients.
In practice, we adopt a four-step scheme ($K=4$), illustrated in Fig.~\ref{fig:fdb}.
The AB step predicts the next balanced skip feature from recent fusion directions, while the AM step refines it using the newly estimated direction.
The balanced feature is constructed at the intermediate encoder resolution (Level-3) to balance spatial detail and semantic robustness.
After integration, a lightweight refinement is applied, and the resulting balanced feature is resized and injected into all decoder stages as shared skip guidance, enabling consistent cross-scale information transfer.

\begin{figure*}[htbp]
    \centering
    \includegraphics[width=0.8\linewidth]{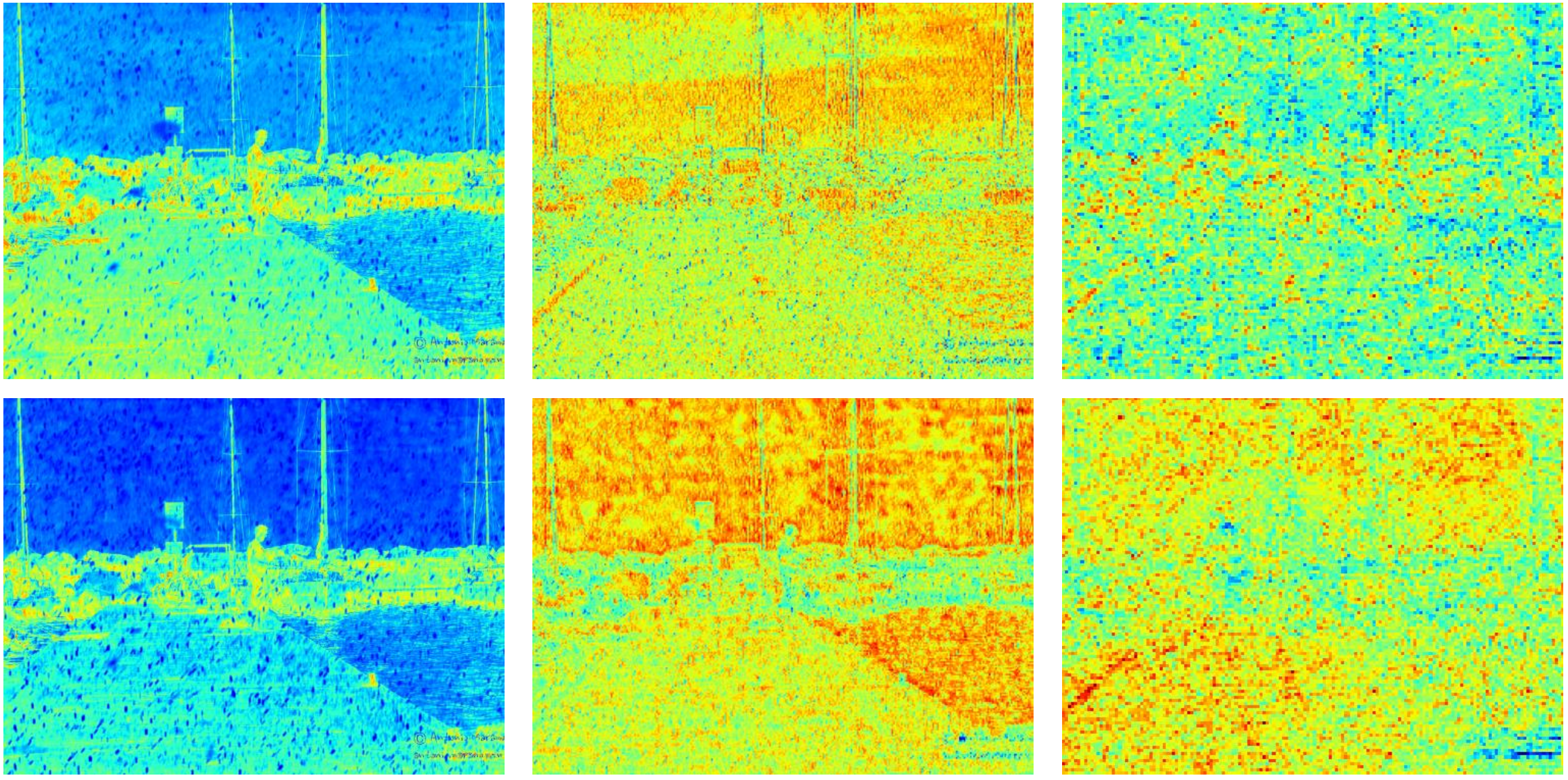}
    \caption{
    Visualization of encoder features (enc level1--level3, top row) and the corresponding
    balanced skip features (balanced level1--level3, bottom row) fed into the decoder.
    Compared to raw encoder features, balanced features exhibit reduced noise and more
    coherent spatial structures, indicating effective cross-scale information integration
    by BMSC.
    }
    \label{fig:bms_visual}
\end{figure*}

Fig.~\ref{fig:bms_visual} compares encoder features and balanced skip features at different scales.
Compared to encoder features, balanced features exhibit reduced noise responses and more coherent spatial structures, suggesting improved cross-scale information integration.

\paragraph{Refinement}
After integration, the balanced skip feature is further refined before being passed to the decoder. In this work, we adopt an efficient attention module to capture global context. The refined skip feature thus serves as a clean and stable multi-scale guidance for decoding under diverse adverse weather conditions. 

To efficiently model long-range dependencies in the refinement stage, we adopt
vHeat~\cite{wang2025building}, a physics-inspired attention mechanism that
reformulates global information aggregation as a heat diffusion process.
Unlike conventional self-attention, which explicitly computes pairwise token
interactions with quadratic complexity, vHeat propagates information smoothly
across the feature map via diffusion, enabling efficient and stable global context modeling for high-resolution features.

\fakeparagraph{Heat diffusion formulation.}
The design of vHeat is motivated by the classical two-dimensional heat equation,
which describes how temperature diffuses over space and time:
\begin{equation}
    \frac{\partial u}{\partial t} = k \left(
    \frac{\partial^2 u}{\partial x^2} +
    \frac{\partial^2 u}{\partial y^2}
    \right),
\end{equation}
where $u(x,y,t)$ denotes the temperature at spatial location $(x,y)$ and time $t$,
and $k$ is the thermal diffusivity.
This equation characterizes a smooth and global propagation process, where information
naturally spreads from each location to the entire spatial domain.
By transforming the heat equation into the frequency domain, the diffusion process
admits a closed-form solution:
\begin{equation}
    \tilde{u}(\omega_x,\omega_y,t)
    = \tilde{f}(\omega_x,\omega_y)\,
    \exp\!\big(-k(\omega_x^2+\omega_y^2)t\big),
\end{equation}
where $\tilde{f}(\cdot)$ is the frequency-domain representation of the input signal.
This formulation reveals that heat diffusion corresponds to a frequency-dependent
attenuation, where different frequency components are modulated according to the
diffusion strength.

\fakeparagraph{Heat Conduction Operator (HCO).}
Building upon this observation, the Heat Conduction Operator (HCO)
refine feature maps in a fully differentiable manner.
Given an input feature map $U_0 \in \mathbb{R}^{H \times W \times C}$, HCO is defined as
\begin{equation}
    U_t =
    \mathrm{IDCT2D}\!\left(
    \mathrm{DCT2D}(U_0)
    \cdot
    \exp\!\big(-k(\omega_x^2+\omega_y^2)t\big)
    \right),
\end{equation}
where $\mathrm{DCT2D}(\cdot)$ and $\mathrm{IDCT2D}(\cdot)$ denote the 2D Discrete Cosine
Transform and its inverse, respectively.
The DCT provides an efficient approximation of the Fourier transform under Neumann
boundary conditions, making it well suited for image-like feature maps.
In this formulation, global context aggregation is achieved through frequency-domain
diffusion rather than explicit token-to-token interaction.
As a result, HCO maintains a global receptive field while avoiding the quadratic
cost of self-attention, achieving a computational complexity of
$\mathcal{O}(N^{1.5})$ for an $N$-pixel feature map.

\fakeparagraph{Adaptive diffusion and refinement.}
To enable content-aware refinement, the diffusion strength $k$ is not fixed but
predicted dynamically using learnable Frequency Value Embeddings (FVEs).
This allows the model to adaptively control the extent of diffusion according to
the input content, balancing global structure propagation and local detail preservation.

Overall, vHeat provides an efficient and interpretable alternative to self-attention
for the refinement stage.
By modeling feature interactions as a diffusion process, it enables smooth global
information propagation, stable optimization, and scalability to high-resolution
inputs, making it particularly suitable for fine-grained image restoration.

The effectiveness of this refinement design is further validated through ablation studies
in Section~\ref{sec:Ablation Studies}, where we compare different refinement strategies and key architectural components. 

\subsection{Uncertainty-Aware Refinement Head (URH) }

Many restoration frameworks generate a coarse prediction followed by a refinement head~\cite{zamir2022restormer,sun2024restoring,wang2025modem,wu2025beyond}. 
However, under severe degradations, coarse outputs often exhibit blurred boundaries and inconsistencies, limiting refinement effectiveness~\cite{qin2019basnet}. 
Moreover, most refinement modules are shallow and single-scale, making it difficult to handle large artifacts or ambiguous structures, and treating refinement as deterministic often leads to over-confident and unstable predictions.
To address these issues, we propose URH, which performs fine-grained correction with uncertainty modeling. 
URH adopts a compact four-stage U-Net with GDTB blocks to enhance multi-scale details, while retaining standard skip connections since cross-scale fusion is already handled by earlier stages.

URH produces a probabilistic output via two parallel heads that predict the per-pixel mean $\mu(x)$ and variance $\sigma^2(x)$. 
Rather than serving as a strictly calibrated uncertainty estimate, the predicted variance acts as a task-driven signal that reflects the relative difficulty of restoration across spatial regions. 
In particular, severely degraded areas (e.g., heavy snow or haze) tend to exhibit higher variance, while clean or well-observed regions show lower variance. 
This design enables uncertainty-aware refinement, where the variance highlights ambiguous regions and modulates the refinement process. 
As a result, the model reduces over-confident predictions in difficult areas and achieves more stable and robust restoration under heterogeneous degradations.

\paragraph{Brightness-Aware Energy Loss}

Most existing adverse-weather restoration methods rely on point-wise $\ell_1$ or $\ell_2$ losses,
which work well in lightly degraded regions but often produce over-confident predictions in severely corrupted or structurally ambiguous areas.
This issue becomes more pronounced when handling diverse weather conditions and degradation levels,
as the resulting ambiguity and uncertainty increase during restoration.
Since adverse-weather restoration is inherently a dense regression problem with spatially varying ambiguity,
explicitly modeling pixel-wise uncertainty is crucial for robust and stable prediction.
To this end, we supervise URH using the proposed BAE-Loss, inspired by heteroscedastic uncertainty modeling
in dense prediction tasks~\cite{cheng2025urnet}.
URH predicts a per-pixel Gaussian distribution parameterized by a mean image $\mu(x)$ and a variance map.
To evaluate the quality of such probabilistic predictions, we adopt an energy-based scoring rule, which is a strictly proper and non-local metric for multivariate probabilistic forecasts~\cite{gneiting2008assessing}.
Given a ground-truth image $z_n$ and $M$ Monte Carlo samples ${z_{n,i}}{i=1}^{M}$ drawn from $\mathcal{N}(\mu(x_n), \Sigma(x_n))$ (using 1000 Monte Carlo samples), the Energy Score is approximated as
\begin{equation}
\begin{aligned}
\mathcal{L}_{\text{ES}}
=
\frac{1}{N}
\sum_{n=1}^{N}
\Bigg(
&
\frac{1}{M}
\sum_{i=1}^{M}
\|z_{n,i}-z_n\|
\\
&
-
\frac{1}{2(M-1)}
\sum_{i=1}^{M-1}
\|z_{n,i}-z_{n,i+1}\|
\Bigg)
\end{aligned}
\label{eq:es_mc}
\end{equation}
where the distance is computed using a pixel-wise $\ell_1$ norm, which provides stable gradients and is better suited to high-resolution image restoration under heavy degradation.
Owing to its non-local nature, this formulation encourages the predicted distribution to place probability mass near the ground truth rather than matching it at a single point, leading to more robust uncertainty estimation.
In addition to uncertainty modeling, adverse-weather images often exhibit global brightness shifts caused by haze, snow accumulation, or illumination changes.
To alleviate this issue, we incorporate a brightness-aware regression term~\cite{liao2025gt}.
Let $f(x)$ denote the predicted mean image and $y$ the ground truth.
The brightness-aware regression loss is defined as
\begin{equation}
\mathcal{L}_{\text{GT}}(f(x), y)
=
W \, \| f(x) - y \|_1
+
(1-W)\,
\left\|
\frac{\mu_y}{\mu_{f(x)}} \, f(x) - y
\right\|_1 ,
\label{eq:gt_loss}
\end{equation}
where $\mu_{f(x)}$ and $\mu_y$ denote the mean brightness of the predicted image
$f(x)$ and the ground truth $y$, respectively.
The adaptive weight $W \in [0,1]$ is computed based on the brightness discrepancy between $f(x)$ and $y$ using a Bhattacharyya-distance-based measure.
Specifically, we model the brightness statistics of the ground truth and prediction as Gaussian distributions:
\begin{equation}
p \sim \mathcal{N}(\mu_y, \sigma_y^2), \quad
q \sim \mathcal{N}(\mu_{f(x)}, \sigma_{f(x)}^2),
\end{equation}
where $\mu_y$ and $\mu_{f(x)}$ denote the mean brightness of $y$ and $f(x)$, and $\sigma_y^2$ and $\sigma_{f(x)}^2$ denote the corresponding variances.
The Bhattacharyya distance between $p$ and $q$ is computed as:
\begin{equation}
D_B(p \| q) =
\frac{1}{4} \frac{(\mu_y - \mu_{f(x)})^2}{\sigma_y^2 + \sigma_{f(x)}^2}
+ \frac{1}{2} \ln \left(
\frac{\sigma_y^2 + \sigma_{f(x)}^2}{2\sigma_y \sigma_{f(x)}}
\right).
\end{equation}
The adaptive weight $W$ is obtained by clipping $D_B$ to the range $[0,1]$.
This formulation adaptively balances the original prediction and its
brightness-aligned counterpart according to global illumination consistency.
Finally, we combine the energy-based uncertainty loss and the brightness-aware regression term to form the proposed Brightness-Aware Energy Loss:
\begin{equation}
\mathcal{L}_{\text{BAE}}
= (1 - w_{\text{prob}}(t))\,\mathcal{L}_{\text{GT}}
+ w_{\text{prob}}(t)\,\mathcal{L}_{\text{ES}},
\end{equation}
where $w_{\text{prob}}(t)\in[0,1]$ is an annealing weight that gradually increases during training.
This design allows URH to first focus on stable brightness-aware reconstruction and progressively incorporate probabilistic supervision, resulting in sharper refinements and better-calibrated uncertainty under severe and ambiguous degradations.

\subsection{Total Loss}
\label{sec:Loss Functions}

The final training objective further incorporates a correlation loss
$\mathcal{L}_{\text{cor}}$~\cite{ozdenizci2023restoring}.
consistency:
\begin{equation}
\mathcal{L}
= \mathcal{L}_{\text{BAE}} + \mathcal{L}_{\text{cor}} .
\end{equation}
The correlation loss is defined based on the Pearson correlation coefficient between
the restored image $I_{\mathrm{HQ}}$ and the ground truth $I_{\mathrm{GT}}$:
\begin{equation}
\begin{aligned}
\mathcal{L}_{\mathrm{cor}}(I_{\mathrm{HQ}}, I_{\mathrm{GT}})
&=
\frac{1}{2}
\left(
1-\rho(I_{\mathrm{HQ}}, I_{\mathrm{GT}})
\right),\\
\rho(I_{\mathrm{HQ}}, I_{\mathrm{GT}})
&=
\frac{
\sum_{i=1}^{N}
(I_{i,\mathrm{HQ}}-\bar{I}_{\mathrm{HQ}})
(I_{i,\mathrm{GT}}-\bar{I}_{\mathrm{GT}})
}{
N\,\sigma(I_{\mathrm{HQ}})
\sigma(I_{\mathrm{GT}})
}.
\end{aligned}
\end{equation}
This loss encourages the restored image to preserve global structural and intensity
consistency with the ground truth, complementing the pixel-wise supervision and
uncertainty modeling in BAE-Loss.

%% file: Sections/4_Experiments.tex
\section{Experimental Setting}
\subsection{Training Details}

Our model is implemented in PyTorch and trained from scratch on four NVIDIA H100 GPUs
for a total of 300{,}000 iterations.
We adopt a progressive learning strategy with five training stages.
At stage $k$, a patch size $s_k$, a per-GPU mini-batch size $b_k$, and a training length of $T_k$ iterations are used, where
$s_{k\in\{1,\dots,5\}}=\{128,160,256,320,360\}$,
$b_{k\in\{1,\dots,5\}}=\{8,5,2,1,1\}$, and
$T_{k\in\{1,\dots,5\}}=\{92{,}000,84{,}000,56{,}000,36{,}000,32{,}000\}$, with
$\sum_k T_k=300{,}000$.
This schedule progressively increases the effective patch size while reducing the batch
size, enabling the network to learn higher-resolution content without exceeding GPU
memory limits.
We use the AdamW optimizer with an initial learning rate of $3\times10^{-4}$, which is
kept constant for the first 92{,}000 iterations and then decayed to $1\times10^{-6}$ using
a cosine annealing schedule over the remaining iterations.
The main architectural hyperparameters are as follows: the numbers of blocks at the four
encoder–decoder stages are $L_{i\in\{1,2,3,4\}}=\{4,4,6,8\}$, the base channel dimension is
$C=36$, the channel expansion factor in DGFF is $r=2.667$, and the numbers of attention
heads at the four stages are $\{1,2,4,8\}$.
For data augmentation, random horizontal and vertical flips are applied during training.

\subsection{Datasets}

\textbf{Snow100K}~\cite{liu2018desnownet} contains 100K synthetic snowy images generated from clean outdoor scenes with different snow densities and particle sizes. Following common practice, we use 9{,}000 images for training. For testing, we adopt three subsets: Snow100K-S (small-particle snow), Snow100K-L (large-particle snow), and Snow100K-Real (real snowy scenes).

\textbf{Raindrop}~\cite{qian2018attentive} provides 1{,}319 real-world image pairs degraded by adherent raindrops. We use 1{,}069 pairs for training and 249 pairs for testing. This dataset focuses on localized, non-uniform occlusions that obscure important image regions.

\textbf{Outdoor-Rain}~\cite{li2019heavy} consists of 9{,}000 synthetic images with combined rain streaks and fog, simulating complex atmospheric degradations. It complements the above datasets by introducing mixed rain–haze conditions.

During training, we merge Snow100K, Raindrop, and Outdoor-Rain into a unified multi-weather training set that covers both synthetic and real degradations. For evaluation, we report results on Snow100K-S/L, the Raindrop test set, and the Outdoor-Rain Test1 split.

\subsection{Evaluation Metrics}

We adopt two standard full-reference metrics to evaluate restoration quality: Peak Signal-to-Noise Ratio (PSNR) and Structural Similarity Index (SSIM).

\paragraph{Peak Signal-to-Noise Ratio (PSNR)}
PSNR is derived from the Mean Squared Error (MSE) between the restored image $I_{hq}$ and ground truth $I_{gt}$:
\begin{equation}
\text{MSE} = \frac{1}{H W}\sum_{i=1}^{H}\sum_{j=1}^{W} \big(I_{hq}(i,j) - I_{gt}(i,j)\big)^2,
\end{equation}
\begin{equation}
\text{PSNR} = 10 \cdot \log_{10}\left(\frac{MAX^2}{\text{MSE}}\right),
\end{equation}
where $H$ and $W$ are height and width, and $MAX$ is the maximum pixel value (e.g., 255 for 8-bit images). Higher PSNR means smaller pixel-wise error.

\paragraph{Structural Similarity Index (SSIM)}
SSIM measures perceptual similarity in terms of luminance, contrast, and structure:
\begin{equation}
\text{SSIM}(I_{hq}, I_{gt}) = 
\frac{(2\mu_{hq}\mu_{gt} + C_1)(2\sigma_{hq,gt} + C_2)}
{(\mu_{hq}^2 + \mu_{gt}^2 + C_1)(\sigma_{hq}^2 + \sigma_{gt}^2 + C_2)},
\end{equation}
where $\mu_{hq}$, $\mu_{gt}$ are mean intensities, $\sigma_{hq}^2$, $\sigma_{gt}^2$ are variances, and $\sigma_{hq,gt}$ is the covariance between $I_{hq}$ and $I_{gt}$. $C_1$ and $C_2$ are small constants for numerical stability. SSIM ranges from $-1$ to $1$, with larger values indicating better structural similarity.

\paragraph{Learned Perceptual Image Patch Similarity (LPIPS)}
LPIPS~\cite{zhang2018unreasonable} is a learned perceptual metric that aligns image similarity with human visual perception more closely than traditional pixel-wise measures. Instead of computing differences in the RGB space, LPIPS evaluates perceptual similarity by measuring distances between deep feature representations extracted from a fixed pretrained convolutional network, such as AlexNet or VGG. Given a restored image $I_{hq}$ and the ground truth $I_{gt}$, both images are forwarded through the network, and let $\phi_l(\cdot)$ denote the activation at the $l$-th layer. Following~\cite{zhang2018unreasonable}, the perceptual distance is computed as a learned weighted feature difference:
\begin{equation}
\begin{aligned}
\mathrm{LPIPS}(I_{hq}, I_{gt})
=&
\sum_{l}
\frac{1}{H_l W_l}
\sum_{h,w}
\Big\|
w_l \odot \Big(
\hat{\phi}_l(I_{hq})_{h,w} \\
&
-
\hat{\phi}_l(I_{gt})_{h,w}
\Big)
\Big\|_2^2 .
\end{aligned}
\end{equation}
where $\hat{\phi}_l(\cdot)$ denotes channel-wise normalized feature maps, $w_l$ are learned per-channel weights, and $H_l$ and $W_l$ are the spatial dimensions at layer $l$. By computing distances in deep feature space, LPIPS captures perceptual differences related to texture, structure, and semantic consistency that are often overlooked by pixel-wise metrics, with lower values indicating higher perceptual similarity.

\paragraph{Q-Align}
Q-Align~\cite{wu2023q} is a no-reference perceptual quality metric proposed to align machine-predicted visual scores with human subjective judgments.
Unlike conventional no-reference IQA models that directly regress continuous scores, Q-Align emulates the human rating process by predicting discrete, text-defined quality levels (e.g., \emph{bad}, \emph{poor}, \emph{fair}, \emph{good}, \emph{excellent}) using a large multimodal model.
During inference, the probabilities of these rating levels are extracted and converted into a final quality score via weighted averaging, analogous to the computation of mean opinion scores (MOS) in subjective studies.
By leveraging semantic reasoning and discrete-level supervision, Q-Align demonstrates strong robustness and cross-dataset generalization, making it particularly suitable for evaluating perceptual quality under complex and previously unseen degradations.

\paragraph{Multi-scale Image Quality (MUSIQ)}
MUSIQ~\cite{ke2021musiq} is a no-reference image quality assessment (NR-IQA) metric that predicts perceptual image quality directly from a single input image, without requiring a reference.
It explicitly models image quality across multiple spatial scales, reflecting the fact that human perception jointly considers local details and global composition.
MUSIQ extracts patch-based representations from the image at multiple resolutions and employs a Transformer to aggregate quality-related cues through self-attention.
By jointly encoding spatial location and scale information, MUSIQ effectively captures both fine-grained distortions and large-scale structural degradations.
The final quality score is obtained by regressing from a global representation that summarizes the multi-scale features.
Higher MUSIQ values indicate better perceptual image quality.



\section{Experiments}

\begin{table*}[t]
  \caption{Quantitative comparison on four adverse-weather benchmarks.
  Values are PSNR (dB) and SSIM. Best results are \textbf{bold}.}
  \label{tab:main_results}
  \vskip 0.1in
  \centering
  \begin{small}
  \setlength{\tabcolsep}{3.0pt}
  \begin{tabular}{l cccccccccc}
    \toprule
    \textbf{Method} &
    \multicolumn{2}{c}{\textbf{Snow100K-S}} &
    \multicolumn{2}{c}{\textbf{Snow100K-L}} &
    \multicolumn{2}{c}{\textbf{Outdoor-Rain}} &
    \multicolumn{2}{c}{\textbf{Raindrop}} &
    \multicolumn{2}{c}{\textbf{Avg.}} \\
    \cmidrule(lr){2-3}
    \cmidrule(lr){4-5}
    \cmidrule(lr){6-7}
    \cmidrule(lr){8-9}
    \cmidrule(lr){10-11}
    &
    \textbf{PSNR} & \textbf{SSIM} &
    \textbf{PSNR} & \textbf{SSIM} &
    \textbf{PSNR} & \textbf{SSIM} &
    \textbf{PSNR} & \textbf{SSIM} &
    \textbf{PSNR} & \textbf{SSIM} \\
    \midrule

    All-in-One~\cite{li2020all} &
    -- & -- & 28.33 & 0.8820 & 24.71 & 0.8980 & 31.12 & 0.9268 & -- & -- \\

    TransWeather~\cite{valanarasu2022transweather} &
    32.51 & 0.9341 & 29.31 & 0.8879 & 28.83 & 0.9000 & 30.17 & 0.9157 & 30.20 & 0.9094 \\

    Restormer~\cite{zamir2022restormer} &
    36.02 & 0.9579 & 30.36 & 0.9068 & 30.03 & 0.9215 & 32.18 & 0.9408 & 32.21 & 0.9317 \\

    Chen et al.~\cite{chen2022learning} &
    34.42 & 0.9469 & 30.22 & 0.9071 & 29.27 & 0.9147 & 31.81 & 0.9309 & 31.43 & 0.9249 \\

    WGWSNet~\cite{zhu2023learning} &
    34.31 & 0.9460 & 30.16 & 0.9007 & 29.32 & 0.9207 & 32.38 & 0.9378 & 31.54 & 0.9263 \\

    WeatherDiff64~\cite{ozdenizci2023restoring} &
    35.83 & 0.9566 & 30.09 & 0.9041 & 29.64 & 0.9312 & 30.71 & 0.9312 & 31.57 & 0.9308 \\

    PromptIR~\cite{potlapalli2023promptir} &
    36.88 & 0.9643 & 31.34 & 0.9200 & 30.80 & 0.9229 & 32.20 & 0.9359 & 32.80 & 0.9357 \\

    DiffUIR-L~\cite{zheng2024selective} &
    -- & -- & 30.64 & 0.9082 & 30.89 & 0.9231 & 31.90 & 0.9368 & -- & -- \\

    Histoformer~\cite{sun2024restoring} &
    37.41 & 0.9656 & 32.16 & 0.9261 & 32.08 & 0.9389 & 33.06 & 0.9441 & 33.68 & 0.9437 \\

    MODEM~\cite{wang2025modem} &
    38.08 & 0.9673 & 32.52 & 0.9292 & 33.10 & 0.9410 & 33.01 & 0.9434 & 34.18 & 0.9452 \\

    HOGformer~\cite{wu2025beyond} &
    37.93 & 0.9685 & 32.41 & 0.9297 & 32.89 & 0.9460 & 32.72 & 0.9452 & 33.99 & 0.9474 \\

    \midrule
    \textbf{Ours} &
    \textbf{38.34} & \textbf{0.9697} &
    \textbf{32.77} & \textbf{0.9324} &
    \textbf{33.40} & \textbf{0.9491} &
    \textbf{33.32} & \textbf{0.9487} &
    \textbf{34.46} & \textbf{0.9500} \\
    \bottomrule
  \end{tabular}
  \end{small}
  \vskip -0.1in
\end{table*}

\begin{table*}[t]
\centering
\caption{Quantitative comparison on adverse-weather removal tasks.}
\label{tab:all_tasks}
\scriptsize
\setlength{\tabcolsep}{2.2pt}
\renewcommand{\arraystretch}{0.88}

\begin{minipage}[t]{0.43\textwidth}
\centering
\caption*{(a) Snow removal}
\begin{tabular}{l cc cc}
\toprule
\textbf{Method} &
\multicolumn{2}{c}{\textbf{Snow-S}} &
\multicolumn{2}{c}{\textbf{Snow-L}} \\
\cmidrule(lr){2-3} \cmidrule(lr){4-5}
& \textbf{PSNR} & \textbf{SSIM} & \textbf{PSNR} & \textbf{SSIM} \\
\midrule
SPANet~\cite{wang2019spatial}      & 29.92 & 0.8260 & 23.70 & 0.7930 \\
JSTASR~\cite{chen2020jstasr}       & 31.40 & 0.9012 & 25.32 & 0.8076 \\
RESCAN~\cite{li2018recurrent}      & 31.51 & 0.9032 & 26.08 & 0.8108 \\
DesnowNet~\cite{liu2018desnownet}  & 32.33 & 0.9500 & 27.17 & 0.8983 \\
DDMSNet~\cite{zhang2021deep}       & 34.34 & 0.9445 & 28.85 & 0.8772 \\
ConvIR~\cite{cui2024revitalizing}  & 37.98 & 0.9686 & 32.11 & 0.9300 \\
MODEM~\cite{wang2025modem}         & 38.08 & 0.9673 & 32.52 & 0.9292 \\
\textbf{Ours}                      & \textbf{38.34} & \textbf{0.9697} & \textbf{32.77} & \textbf{0.9324} \\
\bottomrule
\end{tabular}
\end{minipage}
\hfill
\begin{minipage}[t]{0.27\textwidth}
\centering
\caption*{(b) Rain removal}
\begin{tabular}{l cc}
\toprule
\textbf{Method} & \textbf{PSNR} & \textbf{SSIM} \\
\midrule
CycleGAN~\cite{zhu2017unpaired} & 17.62 & 0.6560 \\
pix2pix~\cite{isola2017image}   & 19.09 & 0.7100 \\
HRGAN~\cite{li2019heavy}        & 21.56 & 0.8550 \\
PCNet~\cite{jiang2021rain}      & 26.19 & 0.9015 \\
MPRNet~\cite{zamir2021multi}    & 28.03 & 0.9192 \\
NAFNet~\cite{chen2022simple}    & 29.59 & 0.9027 \\
MODEM~\cite{wang2025modem}      & 33.10 & 0.9410 \\
\textbf{Ours}                   & \textbf{33.40} & \textbf{0.9491} \\
\bottomrule
\end{tabular}
\end{minipage}
\hfill
\begin{minipage}[t]{0.27\textwidth}
\centering
\caption*{(c) Raindrop removal}
\begin{tabular}{l cc}
\toprule
\textbf{Method} & \textbf{PSNR} & \textbf{SSIM} \\
\midrule
pix2pix~\cite{isola2017image}         & 28.02 & 0.8547 \\
DuRN~\cite{liu2019dual}               & 31.24 & 0.9259 \\
RaindropAttn~\cite{quan2019deep}      & 31.44 & 0.9263 \\
AttentiveGAN~\cite{qian2018attentive} & 31.59 & 0.9170 \\
MAXIM~\cite{tu2022maxim}              & 31.87 & 0.9352 \\
AST~\cite{zhou2024adapt}              & 30.57 & 0.9333 \\
MODEM~\cite{wang2025modem}            & 33.01 & 0.9434 \\
\textbf{Ours}                         & \textbf{33.32} & \textbf{0.9487} \\
\bottomrule
\end{tabular}
\end{minipage}

\end{table*}


Following previous work~\cite{sun2024restoring}, UAR-Net is evaluated on standard benchmarks for adverse-weather restoration~\cite{liu2018desnownet,qian2018attentive,li2019heavy}.








\subsection{Experimental Results and Comparisons}


We evaluate UAR-Net against a wide range of representative adverse-weather image restoration methods, including both unified and task-specific approaches.
Specifically, Table~\ref{tab:main_results} reports comparisons with unified models~\cite{li2020all,valanarasu2022transweather,zamir2022restormer,chen2022learning,zhu2023learning,ozdenizci2023restoring,sun2024restoring,wang2025modem,wu2025beyond}.
In addition, task-specific comparisons on snow, rain, and raindrop removal are presented in Table~\ref{tab:all_tasks}.



As shown in Tables~\ref{tab:main_results} and~\ref{tab:all_tasks}, UAR-Net consistently achieves SOTA performance across all benchmarks.
On unified evaluation (Table~\ref{tab:main_results}), UAR-Net outperforms Histoformer~\cite{sun2024restoring} by an average PSNR margin of +0.78\,dB,
with notable gains on Snow100K-S (+0.93\,dB), Snow100K-L (+0.61\,dB), Outdoor-Rain (+1.32\,dB), and Raindrop (+0.26\,dB),
and further surpasses MODEM~\cite{wang2025modem} and HOGformer~\cite{wu2025beyond} by clear margins.
Task-specific comparisons show consistent improvements as well:
UAR-Net achieves the best PSNR and SSIM on both Snow100K-S and Snow100K-L (Table~\ref{tab:all_tasks}(a)), improves PSNR from 33.10 to 33.40 on Outdoor-Rain (Table~\ref{tab:all_tasks}(b)), and yields a +0.31\,dB PSNR gain on raindrop removal (Table~\ref{tab:all_tasks}(c)).
demonstrating strong robustness to both large-scale and localized adverse-weather degradations. 

\paragraph{Perceptual quality evaluation}
Beyond distortion-based metrics, we further evaluate perceptual quality using both
full-reference and no-reference metrics, including LPIPS~\cite{zhang2018unreasonable}, Q-Align~\cite{wu2023q}, and MUSIQ~\cite{ke2021musiq}.
The results are reported in Table~\ref{tab:perceptual_metrics}.
UAR-Net consistently achieves the lowest LPIPS scores and the highest Q-Align and MUSIQ
scores across all datasets, indicating that our method not only reduces pixel-wise errors
but also produces more perceptually pleasing and natural results.
The consistent improvements on both full-reference and no-reference metrics suggest that
UAR-Net better balances distortion reduction and perceptual fidelity, which is crucial for
real-world adverse-weather restoration.

\begin{table}
\centering
\footnotesize
\setlength{\tabcolsep}{3pt}
\renewcommand{\arraystretch}{1.1}
\caption{Comparison of perceptual metrics. LPIPS is full-reference ($\downarrow$),
while Q-Align and MUSIQ are no-reference metrics ($\uparrow$).}
\label{tab:perceptual_metrics}

\begin{tabular}{c l|cccc}
\toprule
 & \textbf{Method} 
 & \textbf{Snow100K-L} 
 & \textbf{Snow100K-S} 
 & \textbf{Outdoor} 
 & \textbf{Raindrop} \\
\midrule

\multirow{4}{*}{\rotatebox[origin=c]{90}{\textbf{LPIPS}}}
& WeatherDiff~\cite{ozdenizci2023restoring} & 0.0982 & 0.0541 & 0.0887 & \textbf{0.0615} \\
& Histoformer~\cite{sun2024restoring}       & 0.0919 & 0.0445 & 0.0778 & 0.0672 \\
& MODEM~\cite{wang2025modem}                & 0.0880 & 0.0407 & 0.0699 & 0.0650 \\
& \textbf{Ours}                             & \textbf{0.0799} & \textbf{0.0366} & \textbf{0.0650} & 0.0610 \\
\midrule

\multirow{4}{*}{\rotatebox[origin=c]{90}{\textbf{Q-Align}}}
& WeatherDiff~\cite{ozdenizci2023restoring} & 3.4531 & 3.5293 & 3.8691 & 4.0000 \\
& Histoformer~\cite{sun2024restoring}       & 3.7207 & 3.7598 & 4.1445 & 4.0156 \\
& MODEM~\cite{wang2025modem}                & 3.7324 & 3.7695 & 4.1875 & 4.0664 \\
& \textbf{Ours}                             & \textbf{4.0407} & \textbf{4.0177} & \textbf{4.4015} & \textbf{4.2514} \\
\midrule

\multirow{4}{*}{\rotatebox[origin=c]{90}{\textbf{MUSIQ}}}
& WeatherDiff~\cite{ozdenizci2023restoring} & 62.6267 & 63.1278 & 67.4814 & 69.3608 \\
& Histoformer~\cite{sun2024restoring}       & 64.2526 & 64.2581 & 67.7461 & 68.4852 \\
& MODEM~\cite{wang2025modem}                & 64.2438 & 64.2853 & 68.2926 & 69.7925 \\
& \textbf{Ours}                             & \textbf{66.2562} & \textbf{66.3938} & \textbf{70.9819} & \textbf{71.4844} \\
\bottomrule
\end{tabular}

\end{table}

\begin{figure}
    \centering
    \includegraphics[width=\linewidth]{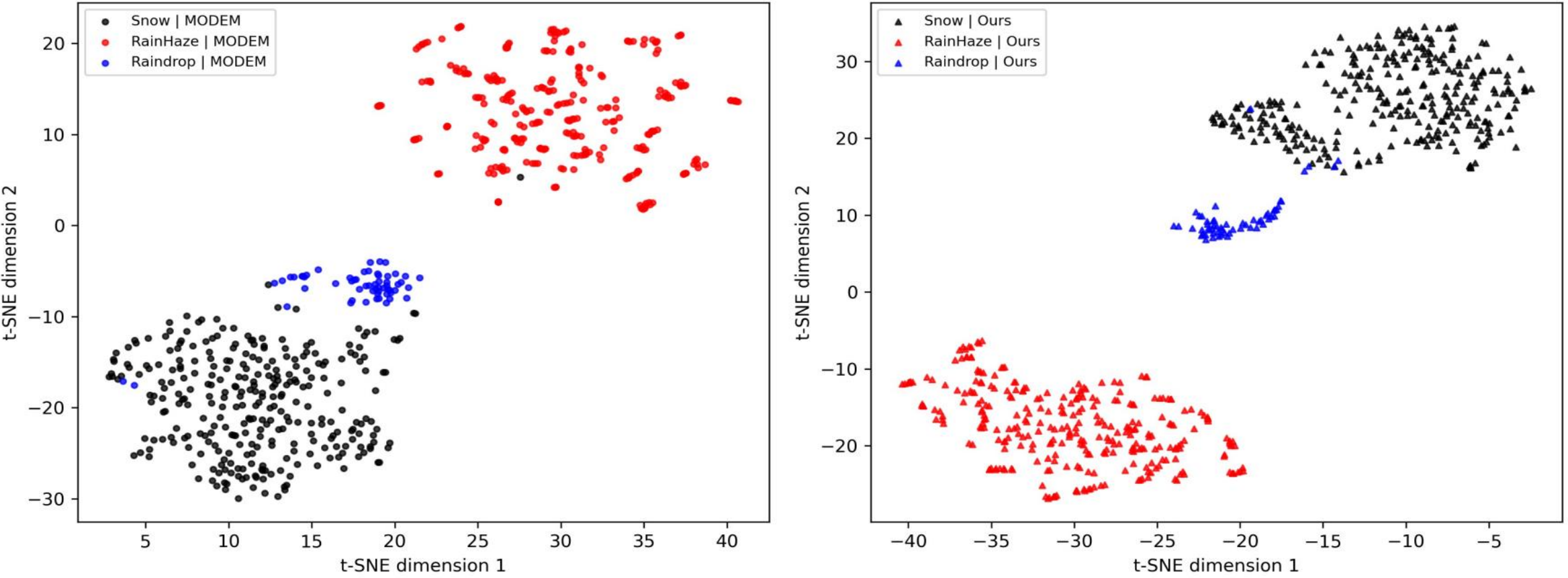}
    \caption{
    T-SNE visualization of features learned by
    MODEM~\cite{wang2025modem} (left) and our method (right).
    Compared with MODEM, our method yields more compact intra-class feature clusters
    and clearer inter-class separation among different adverse-weather conditions,
    indicating improved condition-aware feature disentanglement and representation robustness.
    }
    \label{fig:tsne_modem_compare}
\end{figure}

\paragraph{T-SNE feature visualization}
Fig.~\ref{fig:tsne_modem_compare} presents a t-SNE visualization of the encoder features from MODEM and our method.
MODEM shows scattered feature distributions with noticeable overlap across different weather conditions, indicating limited feature separability.
In contrast, our method yields more compact intra-class clusters and clearer inter-class separation, suggesting more condition-aware and disentangled representations.
This improved feature organization reflects the effectiveness of the gated attention and balanced multi-scale skip design in reducing interference across adverse-weather conditions.

\begin{table}
\vspace{-10pt}
\centering
\footnotesize
\setlength{\tabcolsep}{4pt}
\renewcommand{\arraystretch}{1.0}

\caption{Ablation study on the proposed components.}
\label{tab:ablation_mfs_rhu_bae_avg}
\vspace{-6pt}

\begin{tabular}{c cccc cc}
\toprule
\textbf{Exp.} &
\multicolumn{4}{c}{\textbf{Factors}} &
\multicolumn{2}{c}{\textbf{Avg.}} \\
\cmidrule(lr){2-5} \cmidrule(lr){6-7}
& \textbf{GDTB} & \textbf{BMSC} & \textbf{BAE} & \textbf{URH}
& \textbf{PSNR} & \textbf{SSIM} \\
\midrule
1 & $\times$ & $\times$ & $\times$ & $\times$ & 33.76 & 0.9447 \\
2 & $\checkmark$ & $\times$ & $\times$ & $\times$ & 33.85 & 0.9453 \\
3 & $\checkmark$ & $\checkmark$ & $\times$ & $\times$ & 34.00 & 0.9465 \\
4 & $\checkmark$ & $\checkmark$ & $\checkmark$ & $\times$ & 34.21 & 0.9480 \\
5 & $\checkmark$ & $\checkmark$ & $\checkmark$ & $\checkmark$
  & \textbf{34.46} & \textbf{0.9500} \\
\bottomrule
\end{tabular}

\vspace{-10pt}
\end{table}


\subsection{Ablation Studies}
\label{sec:Ablation Studies}


We present ablation studies on the proposed components and several key design choices of the framework.

\paragraph{Ablation study on the proposed components}
The results are summarized in Table~\ref{tab:ablation_mfs_rhu_bae_avg}.
Starting from the histogram-transformer baseline~\cite{sun2024restoring}, replacing it with GDTB improves the average performance by +0.09\,dB PSNR and +0.0006 SSIM.
Adding BMSC further increases the performance to 34.00\,dB PSNR and 0.9465 SSIM.
Introducing BAE-Loss brings additional gains, reaching 34.21\,dB PSNR and 0.9480 SSIM.
Finally, incorporating URH yields the full model with 34.46\,dB PSNR and 0.9500 SSIM, outperforming the baseline by about +0.70\,dB PSNR and +0.0053 SSIM.
These results demonstrate that GDTB, BMSC, BAE-Loss, and URH contribute positively and complement each other.

\begin{table}[t]
\centering
\caption{Controlled ablations on sinusoidal reweighting and gating. S-S: Snow100K-S, S-L: Snow100K-L.}
\label{tab:sin_gate_ablation}
\scriptsize
\setlength{\tabcolsep}{2.5pt}
\renewcommand{\arraystretch}{1.0}

\caption*{(a) Sinusoidal reweighting (with URH \& BMSC)}
\begin{tabular}{c cc cc cc cc cc}
\toprule
\textbf{Rew.} 
& \multicolumn{2}{c}{S-S}
& \multicolumn{2}{c}{S-L}
& \multicolumn{2}{c}{Outdoor-Rain}
& \multicolumn{2}{c}{Raindrop} \\
\cmidrule(lr){2-3}\cmidrule(lr){4-5}\cmidrule(lr){6-7}\cmidrule(lr){8-9}
& P & S & P & S & P & S & P & S \\
\midrule
$\times$ 
& 37.72 & .9671 & 32.21 & .9268 & 32.39 & .9420 & 32.73 & .9430 \\
$\checkmark$ 
& 37.85 & .9675 & 32.39 & .9282 & 32.33 & .9416 & 32.68 & .9431 \\
\bottomrule
\end{tabular}

\vspace{0.8em}

\caption*{(b) Gating mechanism (with URH)}
\begin{tabular}{c cc cc cc cc cc}
\toprule
\textbf{Gate} 
& \multicolumn{2}{c}{S-S}
& \multicolumn{2}{c}{S-L}
& \multicolumn{2}{c}{Outdoor-Rain}
& \multicolumn{2}{c}{Raindrop} \\
\cmidrule(lr){2-3}\cmidrule(lr){4-5}\cmidrule(lr){6-7}\cmidrule(lr){8-9}
& P & S & P & S & P & S & P & S \\
\midrule
$\times$ 
& 38.18 & .9688 & 32.63 & .9307 & 32.87 & .9454 & 33.06 & .9455 \\
$\checkmark$ 
& 38.34 & .9697 & 32.77 & .9324 & 33.40 & .9491 & 33.32 & .9487 \\
\bottomrule
\end{tabular}

\end{table}

\paragraph{Ablation on sinusoidal reweighting and gating}
Table~\ref{tab:sin_gate_ablation} presents controlled ablations on sinusoidal feature reweighting and the gating mechanism under their respective settings. With URH and BMSC enabled (Table~\ref{tab:sin_gate_ablation}(a)), sinusoidal reweighting yields consistent but modest improvements, providing about +0.05\,dB PSNR gain on average, which indicates its effectiveness in enhancing locality-aware feature interactions. In contrast, when evaluated with URH enabled (Table~\ref{tab:sin_gate_ablation}(b)), the gating mechanism leads to more substantial gains, improving the average PSNR by about +0.27\,dB along with consistent SSIM improvements, highlighting its critical role in adaptively modulating feature responses under diverse weather degradations. Based on these observations, both sinusoidal reweighting and gating are adopted in the final model.

\begin{table}
\vspace{-0.2cm}
\caption{
Ablation on the integration and refinement modules inside BMSC.
}
\label{tab:mfs_integration_refinement}
\centering
\small
\setlength{\tabcolsep}{4pt}
\begin{tabular}{cllcc}
\toprule
\multirow{2}{*}{\textbf{Exp.}} 
& \multicolumn{2}{c}{\textbf{Factors}} 
& \multicolumn{2}{c}{\textbf{Avg.}} \\
\cmidrule(lr){2-3} \cmidrule(lr){4-5}
& \textbf{Integration} 
& \textbf{Refinement} 
& \textbf{PSNR} 
& \textbf{SSIM} \\
\midrule
1 & Avg. & vHeat     & 33.82 & 0.9455 \\
2 & LMF  & CosFormer & 33.93 & 0.9459 \\
3 & LMF  & Nonlocal  & 33.94 & 0.9461 \\
4 & LMF  & vHeat     & \textbf{34.00} & \textbf{0.9465} \\
\bottomrule
\end{tabular}
\vspace{-0.4cm}
\end{table}

\paragraph{Ablation on integration and refinement inside BMSC}
Table~\ref{tab:mfs_integration_refinement} shows an ablation on the integration and refinement
choices inside BMSC, again under a fixed setting without URH and BAE-Loss.
The results indicate that using linear multistep fusion (LMF) instead of simple averaging leads to better average performance: with the same vHeat refinement, LMF achieves about +0.18\,dB higher PSNR and a small SSIM gain.
Under LMF, vHeat also performs slightly better than CosFormer \cite{qin2022cosformer} and Nonlocal \cite{wang2018non}.
Based on these observations, we adopt LMF for integration and vHeat for refinement in our model. 

\paragraph{Balanced feature size in BMSC}
In the BMSC, the balanced feature is the intermediate resolution used by the linear multistep fusion to combine multi-scale encoder features. 
As shown in Table~\ref{tab:balanced_size}, we test three choices for this resolution, $H\times W$, $H/2\times W/2$, and $H/4\times W/4$, under a simplified setting without URH and BAE-Loss (trained with $\ell_1$ loss).
All three options give very similar average PSNR and SSIM, but $H/4\times W/4$ slightly outperforms the others and is also cheaper to compute because of the lower spatial size.
Therefore, we use $H/4\times W/4$ as the default balanced feature size in all subsequent experiments.

\paragraph{The impact of the annealing step in BAE-Loss}
According to Table~\ref{tab:ablation_annealing_uncertainty}, increasing the annealing step from 230k to 300k slightly but consistently improves the average PSNR/SSIM (from 34.16/0.9476 to 34.21/0.9480). This shows that removing the regression loss too early hurts performance, and it is better to keep it almost throughout training. Hence, we set the annealing step to 300k in all subsequent experiments.

\begin{table}[t]
\centering
\small
\caption{Ablation on the balanced feature size
(without URH and BAE-Loss, trained with $\ell_1$ loss).}
\label{tab:balanced_size}
\setlength{\tabcolsep}{4pt}
\renewcommand{\arraystretch}{1.05}
\begin{tabular}{c c cc}
\toprule
\textbf{Exp.} & \textbf{Balanced Size} &
\textbf{PSNR} & \textbf{SSIM} \\
\midrule
1 & $H \times W$     & 33.90 & 0.9458 \\
2 & $H/2 \times W/2$ & 33.90 & 0.9457 \\
3 & $H/4 \times W/4$ & \textbf{34.00} & \textbf{0.9465} \\
\bottomrule
\end{tabular}
\end{table}

\begin{table}[t]
\centering
\small
\caption{Ablation on the annealing step of BAE-Loss.}
\label{tab:ablation_annealing_uncertainty}
\setlength{\tabcolsep}{4pt}
\renewcommand{\arraystretch}{1.05}
\begin{tabular}{c c cc}
\toprule
\textbf{Exp.} & \textbf{Annealing Step} &
\textbf{PSNR} & \textbf{SSIM} \\
\midrule
1 & 230k & 34.12 & 0.9469 \\
2 & 250k & 34.14 & 0.9471 \\
3 & 270k & 34.17 & 0.9478 \\
4 & 300k & \textbf{34.21} & \textbf{0.9480} \\
\bottomrule
\end{tabular}
\end{table}

\subsection{Complexity Analysis}

\begin{table}[t]
\centering
\small
\caption{Model complexity and average restoration performance on a single H100 GPU with $128\times128$ input patches.}
\label{tab:model_complexity}
\setlength{\tabcolsep}{5pt}

\begin{tabular}{lccc}
\toprule
\textbf{Method} &
\textbf{Histoformer}~\cite{sun2024restoring} &
\textbf{MODEM}~\cite{wang2025modem} &
\textbf{Ours} \\
\midrule
FLOPs (GMacs) & 23.27 & 29.08 & \textbf{35.44} \\
Avg. PSNR (dB) & 33.67 & 34.18 & \textbf{34.43} \\
\bottomrule
\end{tabular}
\end{table}

\begin{figure}[htbp]
  \centering
\includegraphics[width=0.9\linewidth]{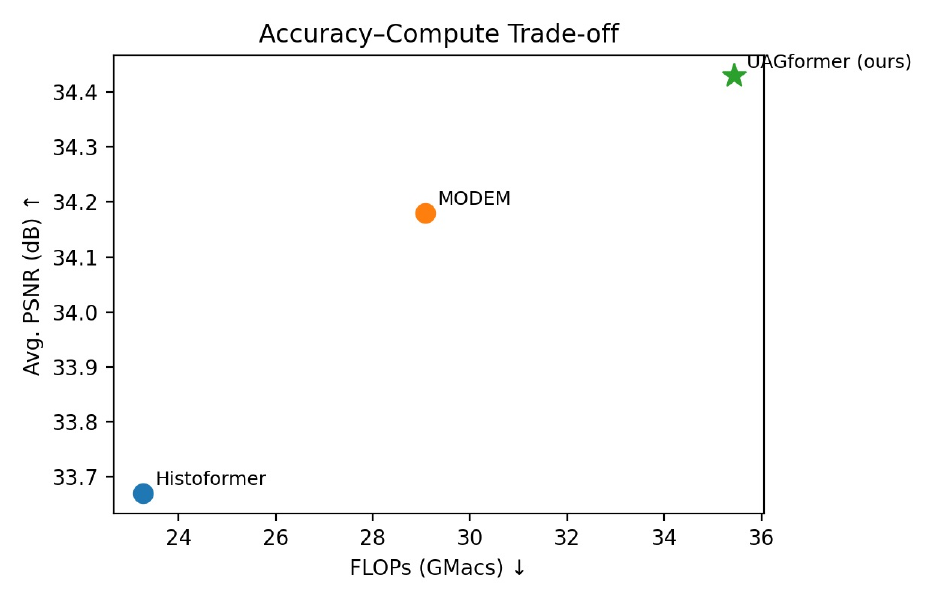}
  \caption{Accuracy--complexity trade-off (Avg. PSNR vs. FLOPs) on $128\times128$ input patches.}
  \label{fig:tradeoff_appendix}
\end{figure}

Table~\ref{tab:model_complexity} reports the computational complexity and average restoration performance of representative unified methods.
UAR-Net achieves the highest average PSNR among all compared models, outperforming MODEM and Histoformer under the same evaluation setting.
Although the compared methods differ in computational cost, UAR-Net consistently delivers superior restoration quality, indicating stronger representation capacity for unified adverse-weather restoration.
As further illustrated in Fig.~\ref{fig:tradeoff_appendix}, UAR-Net lies on a more favorable accuracy--complexity trade-off curve, achieving higher restoration accuracy under comparable computational budgets.

%% file: Sections/5_Conclusion.tex
\section{Conclusion}

In this paper, we proposed UAR-Net, an Uncertainty-guided Adverse-
weather Restoration Network. It integrates GDTB, BMS, and URH to better handle multi-scale structures and residual artifacts, and employs BAE-Loss to jointly learn accurate reconstructions and pixel-wise uncertainty.
Extensive experiments demonstrate SOTA PSNR/SSIM across multiple adverse-weather benchmarks.